%% file: main.tex
\documentclass{article}
\usepackage{ijcai19}

\usepackage{times}
\usepackage{soul}
\usepackage{url}
\usepackage{latexsym}
\usepackage[utf8]{inputenc}
\usepackage[small]{caption}
\usepackage{multicol}
\usepackage{multirow}
\usepackage{graphicx}
\usepackage{amsmath}
\usepackage{booktabs}
\usepackage{algorithm}
\usepackage{algorithmic}
\begin{document}





\title{Earlier Attention? Aspect-Aware LSTM for Aspect-Based Sentiment Analysis}

\author{
Bowen Xing$^1$\footnotemark[1]
\and
Lejian Liao$^1$  \and
 Dandan Song$^{1}$ \footnotemark[2] \and
Jingang Wang $^2$  \and \\
Fuzhen Zhang$^2$   \and
Zhongyuan Wang$^2$ \And
Heyan Huang$^1$
\affiliations
$^1$Lab of High Volume language Information Processing \& Cloud Computing\\
Beijing Lab of Intelligent Information Technology\\
School of Computer Science \& Technology, Beijing Institute of Technology\\
$^2$Meituan-Dianping Group
\emails
\{xingbowen,liaolj,sdd,hhy63\}@bit.edu.cn ,
\{wangjingang02,zhangfuzheng,wangzhongyuan02\}@meituan.com,
}
\maketitle
\renewcommand{\thefootnote}{\fnsymbol{footnote}}
\footnotetext[1]{This work was partially done during Bowen's internship at Meituan-Dianping Group.}
\footnotetext[2]{Corresponding author.}

\input{abstract}
\input{introduction}
\input{relatedwork}
\input{AA-LSTM}
\input{experiment}
\input{resultandconclusion}
\input{acknowledge}
\bibliographystyle{named}
\bibliography{ref}

\end{document}

%% file: abstract.tex
\begin{abstract}
Aspect-based sentiment analysis (ABSA) aims to predict fine-grained sentiments of comments with respect to given aspect terms or categories.
In previous ABSA methods, the importance of aspect has been realized and verified.
Most existing LSTM-based models take aspect into account via the attention mechanism, where the attention weights are calculated after the context is modeled in the form of contextual vectors.
However, aspect-related information may be already discarded and aspect-irrelevant information may be retained in classic LSTM cells in the context modeling process, which can be improved to generate more effective context representations. 
This paper proposes a novel variant of LSTM, termed as aspect-aware LSTM (AA-LSTM), which incorporates aspect information into LSTM cells in the context modeling stage before the attention mechanism. 
Therefore, our AA-LSTM can dynamically produce aspect-aware contextual representations. 
We experiment with several representative LSTM-based models by replacing the classic LSTM cells with the AA-LSTM cells.
Experimental results on SemEval-2014 Datasets demonstrate the effectiveness of AA-LSTM.
\end{abstract}

%% file: introduction.tex
\section{Introduction}

 With increasing numbers of comments on the Internet, sentiment analysis is attracting interests from both research and industry. Aspect-based sentiment analysis is a fundamental and challenging task in sentiment analysis, which aims to infer the sentiment polarity of sentences with respect to given aspects.
For example, \textit{``Great salad but the soup tastes bad".}
 It's obvious that the opinion over the \textit{`salad'} is positive while the opinion over the \textit{`soup'} is negative.
 In this case, aspects are included in the comments, and predicting aspect sentiment polarities of this kind of comments is termed aspect term sentiment analysis (ATSA) or target sentiment analysis (TSA).
 There is another case where the aspect is not explicitly included in the comment.
 For example, \textit{``Although the dinner is expensive, waiters are so warm-hearted!"}. We can observe that there are two aspect categories mentioned in this comment: price and service with completely opposite sentiment polarities.
 Predicting aspect sentiment polarities of this kind of comments is termed aspect category sentiment analysis (ACSA), and the aspect categories usually belong to a predefined set.
 In this paper, we collectively refer aspect category, aspect term/target as aspect. And our goal is aspect-based sentiment analysis (ABSA) including ATSA and ACSA.

 As deep learning have been successfully exploited in various NLP tasks \cite{recall,MachineReading,KGDE,IAD}, many neural networks have been applied to ABSA. With the ability of handling long-term dependencies, Long Short-Term Memory neural network (LSTM) \cite{LSTM} is widely used for context modeling in ABSA, and
 many recent best performing ABSA methods are based on LSTM because of its significant performance \cite{TDLSTM,ATAE,tencent,IAN,IAD,lixin}.
 Current mainstream LSTM-based ABSA models adopt LSTM to model the context, obtaining hidden state vectors for each token in the input sequence.
 After obtaining contextual vector representations, they utilize attention mechanism to produce the attention weight vector.

Recent well performing LSTM-based ABSA models can be divided into three categories according to their way of modeling context:
 (1) ``Attention-based LSTM with aspect embedding (ATAE-LSTM)" \cite{ATAE} and ``modeling inter-aspect dependencies by LSTM (IAD-LSTM)" \cite{IAD} model the context and aspect together via concatenating the aspect vector to the word embeddings of context words in the embedding layer.
 (2) ``Interactive attention networks (IAN)" \cite{IAN} and ``aspect fusion LSTM (AF-LSTM)" \cite{Fusion} model the context alone and utilize the aspect to compute context's attention vector in the attention mechanism.
 (3) ``Recurrent attention network on memory (RAM)" \cite{tencent} introduces relative position information of context words and the given target into their hidden state vectors.


The first category conducts simple joint modeling of contexts and aspects.
In the secondary category, on behalf of most of the existing LSTM-based methods, the models only use context words as input when modeling the context, so they get the same context hidden states vectors when analysing comments containing multiple aspects.
The second category models the context separately while utilizing the aspect information in context's attention calculation.
 And the method in the third category additionally multiplies a relative position weight.
However, no aspect information is considered in the LSTM cells of all the above methods.
Therefore, after context modeling, their hidden state vectors contain the information that is important to the ``overall" comment semantics.
This is determined by the functionality of classic LSTM.
It retains important information and filters out useless information at the sentence-level semantic, in the hidden states corresponding to every context word.


In contrast, for the aspect-based sentiment analysis task, we think the context modeling should be aspect-aware.
 For a specific aspect, on one hand, some of the semantic information of the whole sentence is useless.
 These aspect irrelevant information would adversely harm the final sentiment representation, especially in the situation where multiple aspects exist in one comment.
 This is because when LSTM encounters an important token for the overall sentence semantics, this token's information is retained in every follow-up hidden state.
 Consequently, even if a good attention vector is produced via the attention mechanism, these hidden state vectors also contain useless information which is magnified to some extent.
 On the other hand, information that is important to the aspect may be not sufficiently kept in hidden states because of their small contribution to the overall semantic information of the sentence.

 We take two typical examples to illustrate the two issues.
 First, \textit{``The salad is so delicious but the soup is unsatisfied.''}.
 There are two aspects (\textit{salad} and \textit{soup}) of opposite sentiment polarity.
 When judging the sentiment polarity of the \textit{`soup'}, the word `delicious' which modifies \textit{`salad'} is also important to the sentence-level semantics of the whole comment, and its information is preserved in the hidden states vectors of subsequent context words, including `unsatisfied'.
 So even if `unsatisfied' is assigned a large weight in the attention vector, the information of `delicious' will still be integrated into the final context representation and enlarged.
 Second, \textit{``Pizza is wonderful compared to the last time we enjoyed at another place, and the beef is not bad, by the way.''}
 Obviously, this sentence is mainly about pizza so classic LSTM will retain a lot of information that modifies \textit{`pizza'} when modeling context. But when judging the polarity of beef, because traditional LSTM does not know the aspect is \textit{`beef'}, much-retained \textit{`pizza'} information causes that the information of \textit{`beef'} is not valued enough in hidden state vectors.
 We define the above issues as the aspect-unaware problem in the context modeling process of current methods. To the best of our knowledge, this is the first time to propose this problem.


 In this paper, we propose a novel LSTM variant termed aspect-aware LSTM (AA-LSTM) to introduce the aspect into context modeling process. In every time step, on one hand, the aspect vector can select key information in the context according to the aspect and keep the important information in context words' hidden states. On the other hand, the vector formed aspect information can influent the process of context modeling and filter useless information for the given aspect. So AA-LSTM can generate more effective context hidden states based on the given aspect. This can be seen as an earlier attention operation on context words. It is worth mentioning that though our AA-LSTM model takes the aspect as input, it does not actually fuse the aspect vector into the representation of the context, but only utilize the aspect to influence the process of modeling context via controlling information flow.

 The main contributions of our work can be summarized as follows:
\begin{itemize}
 \item We propose a novel LSTM variant termed as aspect-aware LSTM (AA-LSTM) to introduce the aspect into the process of modeling context.
 \item
 Considering that the aspect is the core information in this task, we fully exploit its potential by introducing it into the LSTM cells. We design three aspect gates to introduce the aspect into the input gate, forget gate and output gate in classic LSTM.
 AA-LSTM can utilize aspect to improve the information flow and then generate more effective aspect-specific context representation.
 \item We apply our proposed AA-LSTM to several representative LSTM-based models, and the experimental results on the benchmark datasets demonstrate the validity and generalization of our proposed AA-LSTM.

\end{itemize}

%% file: relatedwork.tex
\section{Related Work}
In this section, we survey some representative studies in the aspect-based sentiment analysis (ABSA). 
ABSA is the task of predicting the sentiment polarity of a comment with respect to a set of aspects terms or categories included in the context. 
The biggest challenge faced by ABSA is how to effectively represent the aspect-specific sentiment information of the comment \cite{commonsense}.
Although some traditional methods for target sentiment analysis also achieve promising results, they are labor intensive because they have mostly focused on feature engineering or massive extra linguistic resources \cite{NRC2,DCU}.

As deep learning achieved breakthrough success in representation learning, many recent works utilized deep neural networks to automatically extract features and generate the context embedding which is a dense vector formed representation of the comment.

Since the attention mechanism was first introduced to the NLP field \cite{NMT}, many sequence-based approaches utilize it to generate more aspect-specific final representations.
Attention mechanism in ABSA takes aspect information (usually aspect embedding) and the hidden states of every context word (generated by context modeling) as input and produces a probability distribution in which important parts of the context will be assigned bigger weights according to the aspect information. 

There are some CNN-based \cite{cnn} and memory networks (MNs)-based models for context modeling \cite{DMN,Dm,tsmn}.
\cite{Dm} model dyadic interactions between aspect and sentence using neural tensor layers and associative layers with rich compositional operators.
\cite{tsmn} argue that for the case where several sentences are the same except for different targets, relying attention mechanism alone is insufficient. It designed several memory networks having their own characters to solve the problem.

In particular, LSTM networks are widely used in context modeling because of its advantages for sequence modeling \cite{TDLSTM,IAN,IAD,ATAE,Fusion,commonsense,zhangyue,aaai2017}. 
We divide recent well-performing methods into three categories according to the process of modeling context:
First, modeling the context and aspect via concatenating the aspect vector to the word embeddings of context words in the embedding layer.
\cite{ATAE} firstly propose aspect embedding, and their ATAE-LSTM learns to attend to different parts of the context according to the aspect embedding.
Although IAD-LSTM \cite{IAD} model inter-dependencies between multiple aspects of one comment through LSTM after getting the final representation of the context, it is consistent with ATAE-LSTM \cite{ATAE} in the way of context modeling.

Second, modeling the context alone and utilizing the aspect to compute context's attention vector in the attention mechanism.
The main difference among this category of models is the calculation method of the attention mechanism.
\cite{IAN} propose an interactive attention network (IAN) which models targets and contexts separately. Then it learns the interactions between the context and target in attention mechanism utilizing the averages of context's hidden states and target's hidden states.
\cite{Fusion} propose Aspect Fusion LSTM (AF-LSTM) model with a novel association layer after LSTM to model word-aspect relation utilizing circular convolution and circular correlation. 

Third, introducing relative position information of the given target and context words to the hidden state vectors of context words. RAM \cite{tencent} realizes that the hidden state vector of a word will be assigned a larger weight if it is closer to the target through a relative location vector. This operation is conducted before their recurrent attention layer consisting of GRU cells.

Unlike all the above methods, we propose to introduce he aspect information into the process of context modeling. Our proposed AA-LSTM introduces the aspect into the LSTM cells to control information flow. AA-LSTM can not only select key information in the context according to the aspect and keep the important information in context words' hidden state vectors, but also filter useless information for the given aspect. Then AA-LSTM can generate more effective aspect-specific context hidden state vectors.


%% file: AA-LSTM.tex
\section{Aspect-Aware LSTM}
\begin{figure}[!htb]
 \centering
 \includegraphics[scale = 0.5]{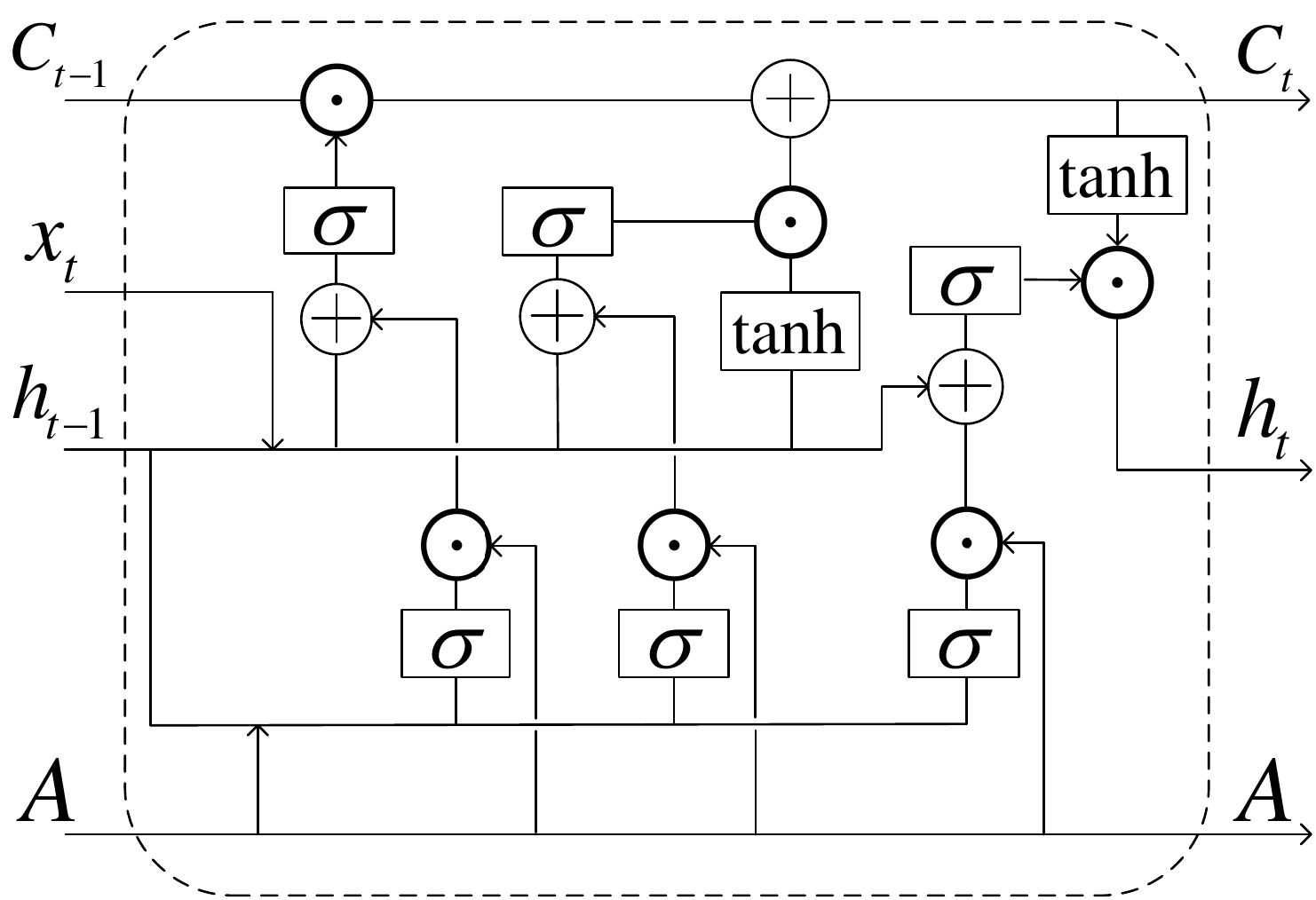}
 \caption{The overall  architecture of AA-LSTM network}
 \label{lstm}
\end{figure}

In this section we describe our proposed aspect-aware LSTM (AA-LSTM) in detail.
Classic LSTM contains three gates (input gate, forget gate and output gate) to control the information flow.
We argue that aspect information should be considered into LSTM cells to improve the information flow. It is intuitive that in every time step the degree that aspect is integrated into the three gates of classic LSTM should be different.
Therefore, we incorporate aspect vector into classic LSTM cells and design three aspect gates to control how much the aspect vector is imported into the input gate, forget gate and output gate respectively.
In this way, we can utilize the previous hidden state and the aspect itself to control how much the aspect is imported in the three gates of classic LSTM.
Figure \ref{lstm} illustrates the architecture of the AA-LSTM network and it can be formalized as follows:
\begin{align}
a_i &= \sigma(W_{ai}\,[A, h_{t-1}] + b_{ai}) \\
I_t &= \sigma(W_I\,[x_t, h_{t-1}] + a_i \odot A + b_I) \\
a_f &= \sigma(W_{af}\,[A, h_{t-1}] + b_{af}) \\
f_t &= \sigma(W_f\,[x_t, h_{t-1}] + a_f \odot A + b_f) \\
\widetilde{C_t} &= tanh(W_C\,[x_i, h_{t-1}] + b_C) \\ \label{candidate}
C_t &= f_t \odot C_{t-1} + I_t \odot \widetilde{C_t} \\
a_o &= \sigma(W_{ao}\,[A, h_{t-1}] + b_{ao}) \\
o_t &= \sigma(W_o\,[x_t, h_{t-1}] + a_o \odot A +b_o) \\
h_t &= o_t * tanh({C_t})
\end{align}
where $x_t$ represents the input embedding vector of the context word corresponding to time step $t$, $A$ stands for the aspect vector, $h_{t-1}$ is previous hidden state, $h_t$ is the hidden state of this time step, $\sigma$ and $tanh$ are sigmoid and hyperbolic tangent functions, $\odot$ stands for element-wise multiplication, $W_{ai}$, $W_{af}$, $W_{ao}\in R^{da\times (dc+da)}$ and $W_I$, $W_f$, $W_C$, $W_o\in R^{dc\times 2dc}$ are the weighted matrices, $b_{ai}, b_{af}, b_{ao} \in R^{da}, b_I, b_f, b_C, b_o \in R^{dc}$ are biases and $da, dc$ stand for the aspect vector's dimension and the number of hidden cells at AA-LSTM respectively.
$I_t, f_t, o_t \in R^{dc}$ stand for the input gate, forget gate and output gate respectively.
The input gate controls the extent of updating the information from the current input.
The forget gate is responsible for selecting some information from last cell state.
The output gate controls how much the information in current cell state is output to be the hidden state vector of this time step.
Similarly, $a_i, a_f, a_o \in R^{da}$ stand for the aspect-input gate, aspect-forget gate and aspect-output gate respectively.
The three aspect-based gates determine the extent of integrating the aspect information into the input gate, forget gate and output gate.

Our proposed AA-LSTM takes two strands of inputs: context word embeddings and the aspect vector.
At each time step, the context word entering the AA-LSTM dynamically varies according to the sequence of words in the sentence, while the aspect vector is identical.
Specifically, aspect vector is the representation of the target in TSA, and it is the aspect embedding in ACSA.
Next, we describe the different components of our proposed AA-LSTM in detail.
\subsection{Input Gates}
The input gate $I_t$ controls how much new information can be transferred to the cell state.
While the aspect-input gate $a_i$ controls how much the aspect is transferred to the input gate $I_t$.
The difference between the AA-LSTM and the classical LSTM lies in the weighted aspect vector input of $I_t$.
The weight of aspect vector $a_i$ is computed by $h_{t-1}$ and $A$.
$h_{t-1}$ can be regarded as the previous semantic representation of the partial sentences which has been processed in the past time steps.
Hence, the extent of the aspect's integration into $I_t$ is decided by the previous semantic representation and the aspect vector $A$.
\subsection{Forget Gates}
The forget gate $f_t$ abandons trivial information and retains key information from last cell state $C_{t-1}$.
Similarly, the aspect-input gate $a_f$ controls how much the aspect vector is transferred to the forget gate $f_t$.
The difference between the AA-LSTM and the classical LSTM in $f_t$ is the introduction of weighted aspect vector.
And the weight of aspect vector $a_f$ is computed by $h_{t-1}$ and $A$.
Therefore, the extent of the aspect's integration into $I_t$ is decided by the previous semantic representation and the aspect vector $A$.
\subsection{Candidate Cell and Current Cell}
The candidate cell $\widetilde{C_t}$ represents the alternative input content.
The current cell $C_t$ updates its cell state by selecting important information from last cell state $C_{t-1}$ and $\widetilde{C_t}$.

From Equation \ref{candidate} we can observe that the alternative input content $\widetilde{C_t}$ aspects two strands of inputs: the last hidden state $h_{t-1}$ and the input embedding $x_{t}$ of this time step.
So the hidden states in each time step contain the semantic information of the previous sentence segment.
In the classical LSTM, information that is more important to the overall sentence semantics is more likely to be preserved in the hidden states vectors of the subsequent time steps. However, for some information which is retained because of its contribution to the semantics of the whole sentence, it may be noisy when judging the sentiment polarity of a given aspect.
And the information that is crucial to analyze the given aspect's sentiment may be neglected due to its less contribution to the overall sentence.
As demonstrated in the Introduction section, we define this phenomenon as an aspect-unaware problem in the process of context modeling.

Our proposed AA-LSTM can solve this problem by introducing aspect to the process of modeling context to control the flow of information. Information that is important for predicting the given aspect's sentiment polarity can be preserved in the hidden states vectors.
In addition, as shown in Equation \ref{candidate}, the alternative input content does not include the aspect information.
So the AA-LSTM only utilizes the given aspect to influent the information flow instead of integrating the aspect information into the hidden state vectors.
\subsection{Output Gates}
The output gate $o_t$ controls the extent of the information flow from the current cell state to the hidden state vector of this time step.
Similarly, the aspect-output gate $a_o$ controls the extent of the aspect's influence on the output gate $I_t$.
The difference between our proposed AA-LSTM and the classical LSTM in $o_t$ is the introduction of weighted aspect vector into $o_t$.
And the weight of aspect vector $a_o$ is computed by $h_{t-1}$ and $A$.
Therefore, the degree of how much the aspect information is integrated into $o_t$ is decided by the previous semantic representation and the aspect vector $A$.

%% file: experiment.tex
\section{Experiment}
 In this section, we introduce the tasks, the datasets, the evaluation metric, the models for comparison and the implementation details.
\subsection{Tasks Definition}
We conduct experiments on two subtasks of aspect sentiment analysis: aspect term sentiment analysis (ATSA) or target sentiment analysis (TSA) and aspect category sentiment analysis (ACSA).
The former infers sentiment polarities of given target entities contained in the context.
The latter infers sentiment polarities of generic aspects such as \textit{`service'} or \textit{`food'} which may or may not be found in the context, and the aspects belong to a predefined set.
In this paper, these two kinds of tasks are both considered and they are collectively named as aspect-based sentiment analysis (ABSA).
\subsection{Datasets}
We experiment on SemEval 2014 \cite{Semeval2014} task 4 datasets which consist of \textit{laptop} and \textit{restaurant} reviews and are widely used benchmarks in many previous works \cite{TDLSTM,ATAE,IAN,tencent,IAD,Fusion,tsmn}.
We remove the reviews having no aspect or the aspects with sentiment polarity of ``conflict''. The dataset we used consists of reviews with at least one aspect labeled with sentiment polarities of \textit{positive, neutral} and \textit{negative}.
For ATSA, we adopt Laptop and Restaurant datasets; And for ACSA, we adopt the Restaurant dataset.
20\% of the training data is used as the development set.
Full statistics of SemEval 2014 task 4 datasets are given in Table 1.
\begin{table}[!hbtp]
\centering
\label{dataset}
\begin{tabular}{|l|l|r|r|r|}
\hline
Task                  & Dataset          & Pos  & Neg & Neu \\ \hline
\multirow{4}{*}{ATSA} & Restaurant Train & 2164 & 807 & 637 \\ \cline{2-5}
                      & Restaurant Test  & 728  & 196 & 196 \\ \cline{2-5}
                      & Laptop Train     & 994  & 870 & 464 \\ \cline{2-5}
                      & Laptop Test      & 341  & 128 & 169 \\ \hline
\multirow{2}{*}{ACSA} & Restaurant Train & 2179 & 839 & 500 \\ \cline{2-5}
                      & Restaurant Test  & 657  & 222 & 94  \\ \hline
\end{tabular}
\caption{Statistic of all datasets}
\end{table}
\subsection{Evaluation Metric}
Since the two tasks are both multi-class classification tasks, we adopt F1-Macro as our evaluation measure.
And there are some other methods that use strict accuracy (Acc) \cite{ATAE,IAN,tencent,IAD} for evaluation, which measures the percentage of correctly predicted samples in all samples. Therefore, we use these two metrics (F1-Macro and Acc) to evaluate the models' performances.

Generally, higher Acc can verify the effectiveness of the system though it biases towards the majority class, and F1-Macro provides more indicative information because the task is a multi-class problem.
\subsection{Models for Comparison}
In order to verify the advantages of our proposed AA-LSTM compared to classic LSTM, we choose some representative LSTM-based models to replace their original LSTM with our proposed AA-LSTM.

In Introduction and Related Work sections, we have divide recent well-performing methods into three categories according to their processes of modeling context.
In order to prove the generalization ability of our model, we select a representative model from each of these categories for experiments.
 We choose ATAE-LSTM, IAN, and RAM as the representatives of the three categories of models because their architectures are novel and they are taken as comparative methods in many works. We also compare our model with the baseline LSTM model. We introduce them in detail as follows:
\paragraph{LSTM.} This is the baseline that ignores targets and only models contexts using one LSTM network. The last hidden state is regarded as the final sentiment representation.
\paragraph{ATAE-LSTM.} It concatenates the aspect embedding to the word embeddings of context words and uses aspect embedding to produce the attention vector. For ATSA, we take the average of the embeddings of the target words as the aspect embedding which is concatenated to the word embeddings of the context words.
\paragraph{IAN.} It models context and target separately and selects important information from them via two interactive attention mechanisms. The target and context can have impacts on the generation of their representations interactively and their representations are concatenated as the final aspect-specific sentiment representations. For the ACSA task, we omit the modeling of the target and use the aspect embedding to produce the attention vector of context words.
\paragraph{RAM.} It utilizes relative location to assign weights to original context hidden state vectors and then learns the attention vector in a recurrent attention mechanism consisting of GRU cell. It can only be applied to ATSA. For the consistent of comparison, we replaced the deep bidirectional LSTM in the original RAM with a unidirectional single-layer LSTM.

We also choose two state-of-the-art methods that are Memory Networks-based and LSTM-based respectively:\\
\paragraph{Target-sensitive Memory Network.} \cite{tsmn} construct six target-sensitive memory networks (TMNs) which have their own characteristics to resolve target sensitivity and got some improvement. We choose the NP (hops) and JCI (hops) that perform best on Laptop and Restaurant, respectively.
\paragraph{Inter-Aspect Dependencies LSTM.} \cite{IAD} model aspect-based sentential representations as a sequence to capture the inter-aspect dependencies.

We don't reimplement the above two models and the results are retrieved from their original papers.
\begin{table*}[!thp]
\fontsize{9}{11}\selectfont
\linespread{1}
\setlength{\tabcolsep}{7mm}{
\begin{tabular}{|c|c|c|c|c|c|c|}
\hline
\multirow{2}{*}{Task and Dataset} & \multicolumn{4}{c|}{ATSA}                                     & \multicolumn{2}{c|}{ACSA}       \\ \cline{2-7}
                                  & \multicolumn{2}{c|}{Laptop} & \multicolumn{2}{c|}{Restaurant} & \multicolumn{2}{c|}{Restaurant} \\ \hline
Model         & F1-Macro        & Acc        & F1-Macro          & Acc          & F1-Macro          & Acc          \\ \hline
LSTM          & 59.77        & 65.99        & 61.04          & 75.00          & 70.07          & 81.71          \\
ATAE-LSTM     & 61.28        & 66.93        & 64.47          & 77.41          & 70.15          & 82.12          \\
IAN           & 64.54        & 70.53        & 65.67          & 78.48          & 70.81          & 83.25          \\
RAM           & 67.05        & 71.32        & 65.84          & 78.57          & -              & -              \\ \hline
IAD-LSTM      & -            & 72.5         & -              & 79.0           & -              & -              \\
JCI (hops)    & 67.2         & 71.8         & \bf 68.8       & 78.8           & -              & -              \\
NP (hops)     & 67.8         & 72.4         & 66.0           & 75.7           & -              & -              \\ \hline \hline
AA-LSTM       & 61.45        & 66.93        & 66.24          & 78.21          & \bf 75.00          & 83.45          \\
ATAE-LSTM (AA) & 62.10        & 69.28        & 66.46          & 78.21          & 74.51          & 83.97          \\
IAN (AA)       & 65.62        & 71.94        & 68.71          & \bf 79.29          & 74.43          & \bf 84.69          \\
RAM (AA)       & \bf 68.47        & \bf 73.20        & 68.15          & 78.13          & -              & -              \\ \hline
\end{tabular}}
\caption{Comparisons of all models on three datasets. Last four models are our proposed AA-LSTM models, and the last three models with suffix ``(AA)" is the variants in which the original classic LSTM is replaced with our proposed AA-LSTM. The results of IAD-LSTM, JCI (hops) and NP (hops) are retrieved from the original papers. Best scores are marked in \textbf{bold}.}
\end{table*}
\subsection{Implementation Details}
We implement the models in Tensorflow.
We initialize all word embeddings by Glove \cite{Glove} and out-of-vocabulary words by sampling from the uniform distribution $U(-0.1,0.1)$. Initial values of all weight matrices are sampled from uniform distribution $U(-0.1,0.1)$ and initial values of all biases are zeros.  All embedding dimensions are set to 300 and the batch size is set as 16. We minimize the loss function to train our models using Adam optimizer \cite{Adam} with the learning rate set as 0.001. To avoid over fitting, we adopt the dropout strategy with $p=0.5$ and the coefficient of \textit{L2} normalization in the loss function is set to 0.01.
All models use softmax classifier.

For ACSA, we initialize all aspect embeddings by sampling from the uniform distribution $U(-0.1,0.1)$.
 As for the input aspect vector ($A$) of our proposed AA-LSTM which is replaced with the classic LSTM in the above models, we set it as follows:
\paragraph{Aspect Term Sentiment Analysis.} We use the average of word embeddings of the target words as $A$ except for IAN.
 For IAN, we use the average of the hidden states vectors of target words as $A$.
\paragraph{Aspect Category Sentiment Analysis.} For all models, we use the aspect embedding as $A$.

We implement all models under the same experiment settings to make sure the improvements based on the original models come from the replacement of classic LSTM with our proposed AA-LSTM.

%% file: resultandconclusion.tex
\section{Results and Analysis}
Our experimental results are illustrated in Table 2.
We can observe that our proposed AA-LSTM and its substitution in other models has an overall advantage over classic LSTMs on their corresponding original models. It especially achieves higher F1-Macro which can better illustrate the overall performances of the models in multiple classes as the classes are unbalanced.
On the ATSA task, except for the F1-Macro score on Restaurant, the performances of our variants overpass the performances of the representative state-of-the-art models.
In the implementation of the experiment, the only difference between original models and their variants is the substitution of classic LSTM. As we replace the original LSTM with our AA-LSTM, the performance improvement can demonstrate the pure effectiveness of our AA-LSTM.

Compared with LSTM, AA-LSTM's improvements on macro are up to 7\% and 6\% on Restaurant for ATSA and ACSA respectively.
Like LSTM, AA-LSTM also directly uses the last hidden state vector as the final sentiment representation sent to the classifier.
But because the aspect is introduced into the process of modeling context, the semantics of the last hidden state vector of AA-LSTM is aspect-specific.
In fact, not only in the last hidden layer, but also in all hidden states vectors, the information which is important for determining the emotional polarity of the aspect is kept, and other useless information is filtered, which makes the context modeling result much better than LSTM.
As the classifiers are the same, the reason AA-LSTM performs better than LSTM is that the final sentiment representation of AA-LSTM is more effective.

AA-LSTM's performance even surpassed ATAE-LSTM and exceed all original models on F1-macro for ACSA.
It is worth mentioning that all baselines utilizes the attention mechanism and ATAE-LSTM also models the context and aspect together via concatenating aspect to every word embeddings of context words.
In contrast, AA-LSTM only models the context without any other processing.
This proves that AA-LSTM's result of modeling context is aspect-specific and effective.
This is because the aspect information is used in the modeling process to control the flow of information, retain and filter information, who performs as earlier attention.

ATAE-LSTM (AA)'s performance exceeds ATAE-LSTM and AA-LSTM.
This shows that AA-LSTM can be compatible with other components of ATAE-LSTM, improving the whole model's performance.
ATAE-LSTM represents a category of models that combine the context and aspect together via concatenating the aspect vector to context word embeddings.
So the experimental results verify that although the input embeddings contain aspect information, it doesn't conflict with the aspect information introduced in AA-LSTM.

IAN represents a category of models which encode the context alone and utilize the aspect to compute contexts' attention vector in the attention mechanism.
IAN-LSTM (AA)'s overall performance exceeds IAN and AA-LSTM.
This proves that the hidden states vectors generated by AA-LSTM can collaborate with the attention mechanism to achieve better performance.

RAM utilizes the relative location vector to assign weights to original context word hidden state vectors, and calculates the attention vector via a recurrent attention mechanism which is more complex than other baseline models.
It is worth noting that compared with RAM, RAM (AA) has more improvement than other original models and their variants.
This is because the advantage of AA-LSTM is amplified in RAM.
In RAM (AA), while the tokens closer to the target are assigned larger weights, AA-LSTM keeps more important information about the target in the tokens closer to the target: adjectives, modifying phrases, clauses, etc.
In addition, the context hidden states vectors generated by AA-LSTM and the recurrent mechanism work together to produce more effective final sentiment representation.

\section{Conclusion}
In this paper, we argue that aspect-related information may be discarded and aspect-irrelevant information may be retained in classic LSTM cells.
To address this problem, we propose a novel LSTM variant termed as Aspect-Aware LSTM.
Due to the introduction of the aspect into the process of modeling context, our proposed Aspect-aware LSTM can select important information about the given target and filter out the useless information via information flow control.
Aspect-Aware LSTM can not only generate more effective contextual vectors than classic LSTM, but also be compatible with other modules.

%% file: acknowledge.tex
\section*{Acknowledgements}
The authors would like to thank the anonymous reviewers for their valuable comments.
This work is supported by National Key Research and Development Program of China (Grant No. 2016YFB1000902), National Natural Science Foundation of China (NSFC, Grant Nos. 61866038 and 61751201), and Research Foundation of Beijing Municipal Science and Technology Commissions (No. Z181100008918002).